\documentclass[10pt,twocolumn,letterpaper,colorlinks]{article}

\usepackage{cvpr}
\usepackage{times}
\usepackage{epsfig}
\usepackage{graphicx}
\usepackage{amsmath}
\usepackage{amssymb}
\usepackage{subfig}
\usepackage{url}

\usepackage[breaklinks=true,bookmarks=false]{hyperref}

\cvprfinalcopy 

\def\cvprPaperID{9029} 

\ifcvprfinal\fi
\begin{document}

\title{Graph Structured Network for Image-Text Matching}

\author{Chunxiao Liu$^{1,2}$, Zhendong Mao$^{3,}$\thanks{Zhendong Mao is the corresponding author.}, Tianzhu Zhang$^{3}$, Hongtao Xie$^{3}$, Bin Wang$^{4}$, Yongdong Zhang$^{3}$\\
$^{1}$Institute of Information Engineering, Chinese Academy of Sciences, Beijing, China\\
$^{2}$School of Cyber Security, University of Chinese Academy of Sciences, Beijing, China\\
$^{3}$University of Science and Technology of China, Hefei, China
$^{4}$Xiaomi AI Lab, Beijing, China\\
{\tt\small liuchunxiao@iie.ac.cn, maozhendong2008@gmail.com,}\\
{\tt\small  \{tzzhang, htxie, zhyd73\}@ustc.edu.cn, wangbin11@xiaomi.com}
}

\maketitle
\thispagestyle{empty}

\begin{abstract}
    Image-text matching has received growing interest since it bridges vision and language. The key challenge lies in how to learn correspondence between image and text. Existing works learn coarse correspondence based on object co-occurrence statistics, while failing to learn fine-grained phrase correspondence. In this paper, we present a novel Graph Structured Matching Network (GSMN) to learn fine-grained correspondence. The GSMN explicitly models object, relation and attribute as a structured phrase, which not only allows to learn correspondence of object, relation and attribute separately, but also benefits to learn fine-grained correspondence of structured phrase. This is achieved by node-level matching and structure-level matching. The node-level matching associates each node with its relevant nodes from another modality, where the node can be object, relation or attribute. The associated nodes then jointly infer fine-grained correspondence by fusing neighborhood associations at structure-level matching. Comprehensive experiments show that GSMN outperforms state-of-the-art methods on benchmarks, with relative Recall@1 improvements of nearly 7\% and 2\% on Flickr30K and MSCOCO, respectively. Code will be released at: \url{https://github.com/CrossmodalGroup/GSMN}.
\end{abstract}

\section{Introduction}

Image-text matching is an emerging task that matches instance from one modality with instance from another modality. This enables to bridge vision and language, which has potential to improve the performance of other multimodal applications. The key challenge in image-text matching lies in learning correspondence of image and text, such that can reflect similarity of image-text pairs accurately.

\begin{figure}[!t]
	\setlength{\abovecaptionskip}{0pt} 
	\setlength{\belowcaptionskip}{-10pt} 
	\begin{center}
		\includegraphics[width=1\linewidth]{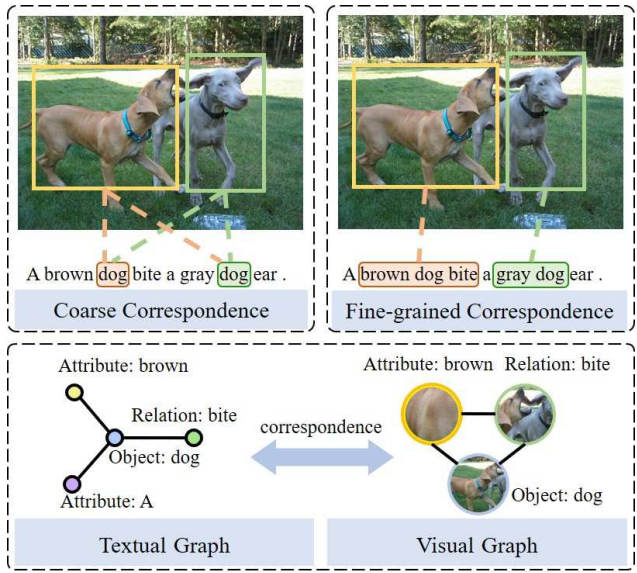}
	\end{center}
	\caption{Illustration of coarse and fine-grained correspondence. In the left figure, the two dogs are coarsely correlated with the word \emph{``dog''}, while neglecting their relation and attribute (bite or being bitten? gray or brown?). In the right figure, the gray and brown dogs are fine-grained correlated with finer textual details, which is achieved by learning phrase correspondence using a graph-based method.}
	\label{fig:long}
	\label{fig:onecol}
\end{figure}

Existing approaches either focus on learning global correspondence or local region-word correspondence. The general framework of global correspondence learning methods is to jointly project the whole image and text into a common latent space, where corresponding image and text can be unified into similar representations. Techniques to common space projection range from designing specific networks \cite{nam2017dual} to adding constraints, such as triplet loss \cite{wang2019learning}, adversarial loss \cite{sarafianos2019adversarial} and classification loss \cite{ Li2017IdentityAwareTM}. Another branch of image-text matching learns local region-word correspondence, which is used to infer the global similarity of image-text pairs. Some researchers focus on learning local correspondence between salient regions and keywords. For example, Ji et al. \cite{ji2019saliency} present to correlate words with partial salient regions detected by a lightweight saliency model, which demands external saliency dataset as a supervision. Recent works discover all possible region-word correspondences. For instance, Lee et al. \cite{lee2018stacked} propose to correlate each word with all the regions with different weights, and vice versa. Following this work, wang et al. \cite{ wang2019position} integrate positional embedding to guide the correspondence learning and Liu et al. \cite{liu2019focus} present to eliminate partial irrelevant words and regions in correspondence learning. 

However, existing works only learn coarse correspondence based on object co-occurrence statistics, while failing to learn fine-grained correspondence of structured object, relation and attribute. As a result, they suffer from two limitations: (1) it is hard to learn correspondences of the relation and attribute as they are overwhelmed by object correspondence. (2) objects are prone to correspond to wrong categories without the guidance of descriptive relation and attribute. As shown in Figure 1, the coarse correspondence will incorrectly correlate the word \emph{``dog''} with all the dogs in the image, while neglecting dogs are with finer details, i.e. brown or gray. By contrast, the fine-grained correspondence explicitly models the object \emph{``dog''}, relation \emph{``bite''} and attribute \emph{``brown''} as a phrase. Therefore, the relation \emph{``bite''} and attribute \emph{``brown''} can also correlate to a specific region, and they will further promote identifying fine-grained phrase \emph{``brown dog bite''}. 

To learn fine-grained correspondence, we propose a Graph Structured Matching Network (GSMN) that explicitly models object, relation and attribute as a phrase, and jointly infer fine-grained correspondence by performing matching on these localized phrases. This unions the correspondence learning of object, relation and attribute in a mutually enforced way. On the one hand, relation correspondence and attribute correspondence can guide the fine-grained object correspondence learning. On the other hand, the fine-grained object correspondence forces the network to learn relation correspondence and attribute correspondence explicitly. Concretely, the proposed network constructs graph for image and text, respectively. The graph node consists of the object, relation and attribute, the graph edge exists if any two nodes interact with each other (e.g. the node of an object will connect with the node of its relations or attributes). Then we perform node-level and structure-level matching on both visual and textual graphs. The node-level matching associates each node with nodes from another modality differentially, which are then propagated to neighborhoods at structure-level matching. The phrase correspondence can be inferred with the guidance of node correspondence. Moreover, the correspondence of object node can be updated as long as its neighboring relation and attribute point to a same object. At last, the updated correspondence is used for predicting the global similarity of image-text pairs, which jointly considers correspondence of all the individual phrases.

The main contributions of this paper are summarized as: (1) We propose a Graph Structured Matching Network that explicitly constructs the graph structure for image and text, and performs matching by learning fine-grained phrase correspondence. To the best of our knowledge, this is the first framework that performs image-text matching on heterogeneous visual and textual graphs. (2) To the best of our knowledge, this is the first work that uses graph convolutional layer to propagate node correspondence, and uses it to infer fine-grained phrase correspondence. (3) We conduct extensive experiments on Flickr30K and MSCOCO, showing our superiority over state-of-the-arts.

\begin{figure*}[t]
	\begin{center}
		\includegraphics[width=\linewidth]{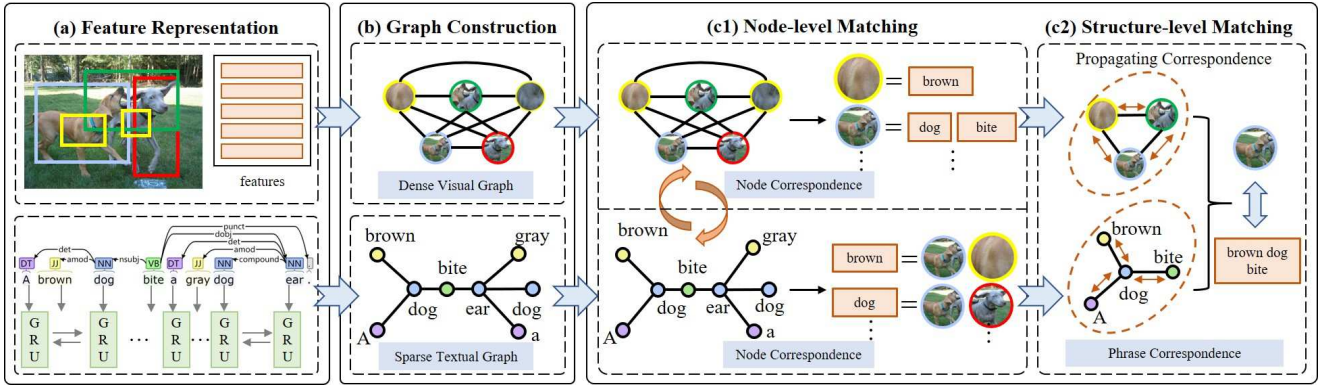}
	\end{center}
	\vspace{-0.3cm}
	\caption{An overview of our approach, which consists of three modules: (a) Feature Extraction: Faster-RCNN \cite{Ren2017Faster} and Stanford CoreNLP \cite{manning-EtAl} are employed to detect salient regions, and parse the semantic dependency, respectively. (b) Graph Construction: The node of graph is object, relation or attribute, the edge exists if any two nodes are semantically dependent. (c1) Node-level Matching: learn correspondence of object, relation and attribute separately. (c2) Structure-level Matching: Propagating the learned correspondence to neighbors to jointly infer fine-grained phrase correspondence.}
	\label{fig:short}
\end{figure*}

\section{Related Work}
Existing works learn correspondence of image and text based on object co-occurence, which is roughly categorized into two types: global correspondence and local correspondence learning methods, where the former learns the correspondence between the whole image and sentence, and the latter learns that between local region and word.

The main goal of global correspondence learning methods \cite{Liu2017LearningAR, Gu2017Look, wang2019learning, wang2019learning, wu2018learning, Mithun2018Webly, zhang2018deep, Eisenschtat2017Linking} is to maximize similarity of matched image-text pairs. A main line of research on this field is to first represent image and text as feature vectors, and then project them into a common space optimized by a ranking loss. Some works focus on designing specific networks. For instance, Liu et al. \cite{Liu2017LearningAR} propose to densely correlate image and text exploiting residual blocks. Gu et al. \cite{Gu2017Look} imagine what the matched instance should look like, and improve the correspondence of target instance to this imagined instance. Some works focus on optimization, Wang et al. \cite{wang2019learning} point out that the correspondence within the same modality should also be preserved while learning correspondence in different modalities. Based on this observation, Wu et al. \cite{wu2018learning} preserve graph structure among neighborhood images or texts. Such global correspondence learning methods cannot learn correspondence of image and text accurately, because primary objects play the dominant role in the global representation of image-text pairs while secondary objects are mostly ignored.

The local correspondence learning methods learn region-word correspondence. Some works focus on learning correspondence of salient objects. Karparthy et al. \cite{KarpathyJF14} make the first attempt by optimizing correspondence of the most similar region-word pairs. Huang et al. \cite{huang2018learning} present to order semantic concepts and composite them to infer correspondence. Similarly, Huang et al. \cite{huang2017instance} propose to recurrently select corresponding region-word pairs. Ji et al. \cite{ji2019saliency} exploit saliency model to localize salient regions, and hence the region-word can be correlated more accurately. A lightweight saliency model is employed using an external saliency dataset as a supervision. The local correspondence policy has also been widely used in other fields \cite{ xu2019multi, liu2016hierarchical}, like \cite{ liu2016hierarchical} that learns distinction and connection among multi-tasks. Another branch of researches \cite{liu2019focus, lee2018stacked, huang2019bi, Ma2015Multimodal,   wang2019position} present to discover all possible region-word correspondence. Ma et al. \cite{Ma2015Multimodal} present to jointly map global image and text, local regions and words into a common space, which can implicitly learn region-word  correspondence. A recent approach SCAN \cite{lee2018stacked} greatly improves the matching performance, which is most relevant to our work. They learn region-word correspondence using attention mechanism, where each region corresponds to multiple words and vice versa. These works learn correspondence based on object co-occurrence, and have achieved much progress in image-text matching. Nonetheless, these only learn coarse correspondence since they mostly rely on correspondence of salient objects, while neglecting the correspondence of relation and attribute is as important as object correspondence. Moreover, the correspondence of relation and attribute can benefit object to correspond to a specific type with a finer detail. By contrast, we explicitly model the image and text as graph structures, and learn fine-grained phrase correspondence. Instead of transforming the image and text as scene graphs using rule-based \cite{yang2019auto,johnson2015image} or classifier-based \cite{schuster2015generating,herzig2018mapping} methods, we only need to identify whether nodes are interact with each other, which avoids the loss of information caused by scene graph generation.

\section{Method}
The overview of our proposed network is illustrated in Figure 2. We first extract features of image and text, and then construct visual and textual graph. Next, the node-level matching learns node correspondence, and propagate to neighbors in structure-level matching, in which the correspondences of object, relation and attribute are fused to infer the fine-grained phrase correspondence.

\subsection{Graph Construction}
\noindent
\textbf{Textual Graph.}
Formally, we seek to construct an undirected sparse graph \begin{math}G_{1}=(V_{1}, E_{1})\end{math} for each text, we use matrix \begin{math}A\end{math} to represent the adjacent matrix of each node, and add self-loops. The edge weight is denoted as a matrix \begin{math}W_{e}\end{math}, which shows the semantic dependency of nodes. 

To construct the textual graph, we first identify the semantic dependency within the text using off-the-shelf Stanford CoreNLP \cite{manning-EtAl}. This can not only parse the object (nouns), relation (verbs) and attribute (adjectives or quantifiers) in a sentence, but also parse their semantic dependencies. For example, given a text \emph{``A brown dog bite a gray dog ear''}, \emph{``A''}, \emph{``brown''} are attributes for the first object \emph{``dog''}, and the \emph{``bite''} is its relation. They are semantically dependent since all of them describe the same object. Based on this observation, we set each word as the graph node, and there exists graph edge between nodes if they are semantically dependent. Then we compute the similarity matrix \begin{math}S\end{math} of word representations \begin{math}u\end{math} as
\begin{equation}
s_{ij}=\frac{ \exp(\lambda u_{i}^{T}u_{j})}{\sum_{j=0}^{m} \exp(\lambda u_{i}^{T}u_{j})}.
\end{equation}
where the \begin{math}s_{ij}\end{math} indicates the similarity between \begin{math}i\end{math}-th and \begin{math}j\end{math}-th node. \begin{math}\lambda\end{math} is a scaling factor. The weight matrix \begin{math}W_{e}\end{math} can be obtained by a Hadamard product between similarity matrix and adjacent matrix, followed by \begin{math}L_{2}\end{math} normalization, i.e.
\begin{equation}W_{e}=\left \| S \circ A\right \|_{2}.\end{equation}

Additionally, we also implement the textual graph as a fully-connected graph. In contrast to sparse graph that employs semantic dependency of words, it can exploit implicit dependencies. We find the sparse and dense graphs are complementary to each other, and can greatly improve the performance, see section 4.2.1.

\noindent
\textbf{Visual Graph.}
To construct the visual graph \begin{math}G_{2}=(V_{2}, E_{2})\end{math}, we represent each image as an undirected fully-connected graph, where the node is set as salient regions detected by Faster-RCNN \cite{Ren2017Faster}, and each node is associated with all the other nodes. Inspired by \cite{norcliffe2018learning} in visual question answering, we use the polar coordinate to model the spatial relation of each image, which disentangles the orientation and distance of pair-wise regions. This can capture both semantic and spatial relationships among different regions, since the relation and attribute are expected to close to object, and the direction information allows to estimate the type of relations. For example, the relations \emph{``on''} and \emph{``under''} show opposite relative position to the object \emph{``desk''}. To get edge weight for this fully-connected graph, we compute polar coordinate \begin{math}(\rho ,\theta)\end{math} based on the centres of the bounding boxes of pair-wise regions, and set the edge weight matrix \begin{math}W_{e}\end{math} as pair-wise polar coordinates.

\subsection{Multimodal Graph Matching}
Given a textual graph \begin{math}G_{1}=(V_{1}, E_{1})\end{math} of a text, and a visual graph \begin{math}G_{2}=(V_{2}, E_{2})\end{math} of an image, our goal is to match two graphs to learn fine-grained correspondence, producing similarity \begin{math}g(G_{1}, G_{2})\end{math} as global similarity of an image-text pair. We define the node representation of textual graph as $ U_{\alpha }\in \mathbb{R}^{m\times d}$, and the node representation of visual graph as $ V_{\beta }\in \mathbb{R}^{n\times d}$. Here, $m$ and $n$ denotes the node number of textual and visual graph, $d$ is the representation dimension. To compute the similarity of these heterogeneous graphs, we first perform node-level matching to associate each node with nodes from another modality graph, i.e. learning node correspondence, and then perform structure-level matching i.e. learning phrase correspondence, by propagating associated nodes to neighbors, which jointly infer fine-grained correspondence of structured object, relation and attribute.

\subsubsection{Node-level Matching}
Each node in the textual and visual graphs will match with nodes from another modality graph to learn node correspondence. We first depict the node-level matching on textual graph in details, and then roughly describe that on visual graph since this operation is symmetric on two kinds of graphs. Concretely, we first compute similarities between visual and textual nodes, denoted as $U_{\alpha }V_{\beta}^{T}$, followed by a softmax function along the visual axis. The similarity value measures how the visual node corresponds to each textual node. Then, we aggregate all the visual nodes as a weighted combination of their feature vectors, where the weight is the computed similarities. This process can be formulated as:
\begin{equation}
C_{t\rightarrow i}={\rm softmax_{\beta}}(\lambda U_{\alpha }V_{\beta }^{T})V_{\beta}.
\end{equation}
where \begin{math}\lambda\end{math} is a scaling factor to focus on matched nodes.

Unlike previous approaches \cite{Fan2018StackedLA, huang2019bi, lee2018stacked} that uses the learned correspondence to compute the global similarity, we present a multi-block module that computes block-wise similarity of the textual node and the aggregated visual node $C_{t\rightarrow i}$. This is computational efficiency and converts the similarity from a scalar into a vector for subsequent operations. Also, this allows different blocks to play different roles in matching. Concretely, we split the $i$-th feature of the textual node and the its corresponding aggregated visual nodes into \begin{math}t\end{math} blocks, represented as \begin{math}[u_{i1},u_{i2},\cdots, u_{it}]\end{math} and \begin{math}[c_{i1},c_{i2},\cdots, c_{it}]\end{math}, respectively. The multi-block similarity is computed within pair-wise blocks. For instance, the similarity in \begin{math}j\end{math}-th blocks is calculated as \begin{math}x_{ij} = cos(u_{ij}, c_{ij})\end{math}. Here, \begin{math}x_{ij}\end{math} is a scalar value, \begin{math}
cos(\cdot)\end{math} denotes cosine similarity. The matching vector of \begin{math}i\end{math}-th textual node can be obtained by concatenating the similarity of all the blocks, that is
\begin{equation}
x_{i}= x_{i1}\left| \right|x_{i2}\left| \right|\cdots \left| \right|x_{it}.
\end{equation}
where ``$\left| \right|$'' indicates concatenation. In this way, each textual node is associated with its matched visual nodes, which will be propagated to its neighbors at structure-level matching to guide neighbors learn fine-grained phrase correspondence.

Symmetrically, when given a visual graph, the node-level matching is proceeded on each visual node. The corresponding textual nodes will be associated differentially 
\begin{equation}
C_{i\rightarrow t}={\rm softmax_{\alpha}}(\lambda V_{\beta }U_{\alpha }^{T})U_{\alpha }
\end{equation}

Then each visual node, together with its associated textual nodes, will be processed by the multi-block module, producing the matching vector $x$.

\subsubsection{Structure-level Matching}
The structure-level matching takes the node-level matching vectors as input, and propagates these vectors to neighbors along with the graph edge. Such a design benefits to learn fine-grained phrase correspondence as neighboring nodes guide that. For example, a sentence \emph{``A brown dog bite a gray dog ear''}, the first \emph{``dog''} will correspond to the visual brown dog in a finer level, because its neighbors \emph{``bite''} and \emph{``brown''} point to the brown dog, and hence the \emph{``dog''} prefer to correlate with the correct dog in the image. To be specific, the matching vector of each node is updated by integrating neighborhood matching vectors using GCN. The GCN layer will apply \begin{math}K\end{math} kernels that learn how to integrate neighborhood matching vectors, formulated as
\begin{equation}
\hat{x}_{i}=\left| \right|_{k=1}^{K}\sigma \left ( \sum_{j\in N_{i}} W_{e}W_{k}x_{j}+b \right ). 
\end{equation}
where \begin{math}N_{i}\end{math} denotes the neighborhood of \begin{math}i\end{math}-th node, \begin{math}W_{e}\end{math} indicates the edge weight depicted in section 3.1, \begin{math}W_{k}\end{math} and \begin{math}b\end{math} are the parameters to be learned of \begin{math}k\end{math}-th kernel. Note that \begin{math}k\end{math} kernels are applied, the output of the spatial convolution is defined as a concatenation over the output of \begin{math}k\end{math} kernels, producing convolved vector that reflects the correspondence of connected nodes. These nodes form the localized phrase. 

The phrase correspondence can be inferred by propagating neighboring node correspondence, which can be used to reason the overall matching score of image-text pair. Here, we feed the convolved vectors into a multi-layer perceptron (MLP) to jointly consider the learned correspondence of all the phrases, and infer the global matching score. This represents how much one structured graph matches another structured graph. This process is formulated as
\begin{eqnarray}
s_{t\rightarrow i}=\frac{1}{n}\sum_{i} W_{s}^{u}(\sigma (W_{h}^{u}\hat{x}_{i}+b_{h}^{u}))+b_{s}^{u}, \\
s_{i\rightarrow t}=\frac{1}{m}\sum_{j} W_{s}^{v}(\sigma (W_{h}^{v}\hat{x}_{j}+b_{h}^{v}))+b_{s}^{v}.
\end{eqnarray}
where \begin{math}Ws,bs\end{math} denote parameters of MLP, which includes two fully-connected layers, the function \begin{math}\sigma(\cdot)\end{math} indicates the tanh activation. Note that we perform structure-level matching on both visual and textual graphs, which can learn phrase correspondence complement to each other. The overall matching score of an image-text pair is computed as the sum of matching score at two directions
\begin{equation}g(G_{1},G_{2})=s_{t\rightarrow i}+s_{i\rightarrow t}.\end{equation}

\subsubsection{Objective Function}
Following previous approaches \cite{liu2019focus,lee2018stacked,Faghri2017VSE,wang2019position}, we employ the triplet loss as the objective function. When using the text \begin{math}T\end{math} as query, we sample its matched images and mismatched images at each mini-batch, which form positive pairs and negative pairs. The similarity in positive pairs should be higher than that in negative pairs by a margin \begin{math}\gamma\end{math}. Analogously, when using the image \begin{math}I\end{math} as query, the negative sample should be a text that mismatches the given query, their similarity relative to positive pairs should also satisfy the above constraints. We focus on optimizing hard negative samples that produce the highest loss, that is \begin{equation}L=\sum_{(I,T)}[\gamma -g(I,T)+g(I,T^{'})]_{+}+[\gamma -g(I,T)+g(I^{'},T)]_{+}.\end{equation}
where \begin{math}I^{'},T^{'}\end{math} are hard negatives, the function \begin{math}[\cdot]_{+}\end{math} is equivalent to \begin{math}max[\cdot,0]\end{math}, and \begin{math}g(\cdot)\end{math} is the global similarity of an image-text pair computed by equation 9.

\subsection{Feature Representation}
\noindent
\textbf{Visual Representation.}
Given an image \begin{math}I\end{math}, we represent its feature as a combination of its \begin{math}n\end{math} salient regions, which are detected by Faster-RCNN pretrained on Visual Genome \cite{krishna2017visual}. The detected regions are feed into pretrained ResNet-101 \cite{He2016DeepRL} to extract features, and then transformed into a \begin{math}d \end{math}-dimensional feature space using a fully connected layer:
\begin{equation}v_{i}=W_{m}[CNN(I_{i})]+b_{m}.\end{equation}
where \begin{math}CNN(\cdot)\end{math} encodes each region within bounding box as a region feature, \begin{math}
W_{m},b_{m}\end{math} are parameters of the fully connected layer that transforms the feature into the common space. These region features form the image representation, denoted as \begin{math} \left [v_{1}, v_{2}, \cdots,v_{n}  \right ]\end{math}.

\noindent
\textbf{Textual Representation.}
Given a text \begin{math}T\end{math} that contains \begin{math}m\end{math} words, we represent its feature as \begin{math} \left [u_{1}, u_{2}, \cdots,u_{m}  \right ]\end{math}, where each word is associated with a feature vector. We first represent each word as a one-hot vector, and then embed it into \begin{math}d\end{math}-dimensional feature space using a Bidirectional Gated Recurrent Unit (BiGRU), which enables to integrated forward and backward contextual information into text embeddings. The representation of  \begin{math}i\end{math}-th word is obtained by averaging the hidden state of forward and backward GRU at \begin{math}i\end{math}-th time step.

\begin{table*}[]
	\centering
	\caption{Image-text matching results on Flickr30K, $'ft'$ and $'fixed'$ are fine-tuning and no fine-tuning. The bests are in bold. }
	\begin{tabular}{|l|c|c|ccc|ccc|c|}
		\hline
		\multicolumn{1}{|c|}{} & \multicolumn{1}{l|}{} & \multicolumn{1}{l|}{} & \multicolumn{3}{c|}{Image-to-Text} & \multicolumn{3}{c|}{Text-to-Image} &  \\ \cline{4-9}
		Method & Image Backbone & Text Backbone & R@1 & R@5 & R@10 & R@1 & R@5 & R@10 & rSum \\ \hline
		m-CNN \cite{Ma2015Multimodal} & fixed VGG-19 & ft CNN & 33.6 & 64.1 & 74.9 & 26.2 & 56.3 & 69.6 & \multicolumn{1}{l|}{324.7} \\
		DSPE \cite{wang2019learning} & fixed VGG-19 & w2v+HGLMM & 40.3 & 68.9 & 79.9 & 29.7 & 60.1 & 72.1 & \multicolumn{1}{l|}{351.0} \\
		VSE++ \cite{Faghri2017VSE} & ft ResNet-152 & ft GRU & 52.9 & 79.1 & 87.2 & 39.6 & 69.6 & 79.5 & 407.9 \\
		TIMAM \cite{sarafianos2019adversarial}  & fixed ResNet-152 & Bert & 53.1 & 78.8 & 87.6 & 42.6 & 71.6 & 81.9 & \multicolumn{1}{l|}{415.6} \\
		DANs \cite{nam2017dual} & ft ResNet-152 & ft LSTM & 55.0 & 81.8 & 89.0 & 39.4 & 69.2 & 79.1 & 413.5 \\
		SCO \cite{huang2018learning} & fixed ResNet-152 & ft LSTM & 55.5 & 82.0 & 89.3 & 41.1 & 70.5 & 80.1 & 418.5 \\
		GXN \cite{Gu2017Look} & ft ResNet-152 & ft GRU & 56.8 & - & 89.6 & 41.5 & - & 80.1 & 268.0 \\
		SCAN \cite{lee2018stacked} & Faster R-CNN & ft Bi-GRU & 67.4 & 90.3 & 95.8 & 48.6 & 77.7 & 85.2 & 465.0 \\
		BFAN \cite{liu2019focus} & Faster R-CNN & ft Bi-GRU & 68.1 & 91.4 & - & 50.8 & 78.4 & - & \multicolumn{1}{l|}{288.7} \\
		PFAN \cite{wang2019position} & Faster R-CNN & ft Bi-GRU & 70.0 & 91.8 & 95.0 & 50.4 & 78.7 & 86.1 & 472.0 \\ \hline
		GSMN (sparse) & Faster R-CNN & ft Bi-GRU & 71.4 & 92.0 & 96.1 & 53.9 & 79.7 & 87.1 & 480.1 \\
		GSMN (dense) & Faster R-CNN & ft Bi-GRU & 72.6 & 93.5 & 96.8 & 53.7 & 80.0 & 87.0 & 483.6 \\
		GSMN (sparse+dense) & Faster R-CNN & ft Bi-GRU & \textbf{76.4} & \textbf{94.3} & \textbf{97.3} & \textbf{57.4} & \textbf{82.3} & \textbf{89.0} & \textbf{496.8} \\ \hline
	\end{tabular}
\end{table*}

\begin{table*}[]
	\centering
	\caption{Image-text matching results on MSCOCO, $'ft'$ and $'fixed'$ are fine-tuning and no fine-tuning. The bests are in bold.}
	\begin{tabular}{|l|c|c|ccc|ccc|c|}
		\hline
		\multicolumn{1}{|c|}{} & \multicolumn{1}{l|}{} & \multicolumn{1}{l|}{} & \multicolumn{3}{c|}{Image-to-Text} & \multicolumn{3}{c|}{Text-to-Image} &  \\ \cline{4-9}
		Method & Image Backbone & Text Backbone & R@1 & R@5 & R@10 & R@1 & R@5 & R@10 & rSum \\ \hline
		m-CNN \cite{Ma2015Multimodal} & fixed VGG-19 & ft CNN & 42.8 & 73.1 & 84.1 & 32.6 & 68.6 & 82.8 & 384.0 \\
		DSPE \cite{wang2019learning} & fixed VGG-19 & w2v+HGLMM & 50.1 & 79.7 & 89.2 & 39.6 & 75.2 & 86.9 & 420.7 \\
		VSE++ \cite{Faghri2017VSE} & ft ResNet-152 & ft GRU & 64.7 & - & 95.9 & 52.0 & - & 92.0 & 304.6 \\
		DPC \cite{zheng2017dual} & \multicolumn{1}{l|}{ft ResNet-152} & \multicolumn{1}{l|}{ft ResNet-152} & 65.5 & 89.8 & 95.5 & 47.1 & 79.9 & 90.0 & 467.8 \\
		GXN \cite{Gu2017Look} & ft ResNet-152 & ft GRU & 68.5 & - & 97.9 & 56.6 & - & 94.5 & 317.5 \\
		SCO \cite{huang2018learning} & fixed ResNet-152 & ft LSTM & 69.9 & 92.9 & 97.5 & 56.7 & 87.5 & 94.8 & 499.3 \\
		SCAN \cite{lee2018stacked} & Faster R-CNN & ft Bi-GRU & 72.7 & 94.8 & 98.4 & 58.8 & 88.4 & 94.8 & 507.9 \\
		BFAN \cite{liu2019focus} & Faster R-CNN & ft Bi-GRU & 74.9 & 95.2 & - & 59.4 & 88.4 & - & 317.9 \\
		PFAN \cite{wang2019position} & Faster R-CNN & ft Bi-GRU & 76.5 & 96.3 & \textbf{99.0} & 61.6 & 89.6 & 95.2 & 518.2 \\ \hline
		GSMN (sparse) & Faster R-CNN & ft Bi-GRU & 76.1 & 95.6 & 98.3 & 60.4 & 88.7 & 95.0 & 514.0 \\
		GSMN (dense) & Faster R-CNN & ft Bi-GRU & 74.7 & 95.3 & 98.2 & 60.3 & 88.5 & 94.6 & 511.6 \\
		GSMN (sparse+dense) & Faster R-CNN & ft Bi-GRU & \textbf{78.4} & \textbf{96.4} & 98.6 & \textbf{63.3} & \textbf{90.1} & \textbf{95.7} & \textbf{522.5} \\ \hline
	\end{tabular}
\end{table*}

\section{Experiment}
\subsection{Dataset and Implementation Details}
To validate the effectiveness of our proposed method, we evaluate it on two most widely used benchmarks, Flickr30K \cite{Plummer2017Flickr30k} and MSCOCO \cite{lin2014microsoft}. Each benchmark contains multiple image-text pairs, where each image is described by five corresponding sentences. Flickr30K collects 31,000 images and 31,000 \begin{math}\times\end{math} 5 = 155,000 sentences in total. Following the settings in previous works \cite{KarpathyJF14},  this benchmark is split into 29,000 training images, 1,000 validation images, and 1,000 testing images. A large-scale benchmark MSCOCO contains 123,287 images and 123,287 \begin{math}\times\end{math} 5 = 616,435 sentences, we use 113,287 images for training, both the validation and testing sets contain 5,000 instances. The evaluation result is calculated on 5-folds of testing images.

The commonly used evaluation metrics for image-text matching are Recall@K (K=1,5,10), denoted as R@1, R@5, and R@10, which depict the percentage of ground truth being retrieved at top 1, 5, 10 results, respectively. The higher Recall@K indicates better performance. Additionally, to show the overall matching performance, we also compute the sum of all the Recall values (rSum) at image-to-text and text-to-image directions, that is 
\begin{equation}
rSum =  \underbrace{R@1 + R@5 + R@10}_{Image\;as\;query} + \underbrace{R@1 + R@5 + R@10}_{Text\;as\;query}.
\end{equation}

As for implementation details, we train the proposed network on training set and validate it at each epoch on validation set, selecting the model with the highest rSum to be test. We train the proposed method on 1 Titan Xp GPU with 30 and 20 epochs for Flickr30K and MSCOCO, respectively. The Adam optimizer is employed with mini batch size 64. The initial learning rate is set as 0.0002 with decaying 10\% every 15 epochs on Flickr30K, and 0.0005 with decaying 10\% every 5 epochs on MSCOCO. We set the dimension of word embeddings as 300, which are then feed into Bi-GRU to get 1024-diemensioanl word representation. As for image feature, each image contains 36 regions that are most salient, and extract 2048-dimensional features for each region. The region feature is then transformed into a 1024-dimensional visual representation by a fully-connected layer. At the structure-level matching, we use one spatial graph convolution layer with 8 kernels, each of which are 32-dimensional. After that, we feed each node in the graph into two fully-connected layers followed by a tanh activation to reason the matching score. The scaling factor $\lambda$ setting is investigated at section 4.2.3. As for optimization, the margin \begin{math}\gamma\end{math} is empirically set as 0.2.

\subsection{Experimental Results}
\subsubsection{Comparisons with state-of-the-arts}
\textbf{Baselines.} we make a comparison with several networks in image-text matching, including (1) typical works m-CNN \cite{Ma2015Multimodal}, DSPE \cite{wang2019learning} and DANs \cite{nam2017dual} that learn global image-text correspondence by designing different network blocks. (2) VSE++ \cite{Faghri2017VSE}, DPC \cite{zheng2017dual} and TIMAM \cite{sarafianos2019adversarial} that learn correspondence using different optimization. (3) SCO \cite{huang2018learning}, GXN \cite{Gu2017Look} that learn region-word correspondence by designing specific networks. (4) state-of-the-art methods SCAN \cite{lee2018stacked}, BFAN \cite{liu2019focus}, PFAN \cite{wang2019position}.
\begin{table}[]
	\centering
	\caption{The ablation study on Flickr30K to investigate the effect of different network structures.}
	\begin{tabular}{|l|cc|cc|}
		\hline
		\multicolumn{1}{|c|}{} & \multicolumn{2}{c|}{Image-to-Text} & \multicolumn{2}{c|}{Text-to-Image} \\ \cline{2-5} 
		Model & R@1 & R@10 & R@1 & R@10 \\ \hline
		GSMN-w/o graph & 63.2 & 94.5 & 48.7 & 84.5 \\
		GSMN-w/o t2i & 64.6 & 93.5 & 45.8 & 82.6 \\
		GSMN-w/o i2t & 67.0 & 95.5 & 52.3 & 86.3 \\
		GSMN-2GCN & 68.4 & 94.8 & 51.5 & 86.0 \\ 
		GSMN-GRU & 71.1 & 95.3 & 50.9 & 85.6 \\
		GSMN-full (sparse)  & 71.4 & 96.1 & 53.9 & 87.1 \\
		GSMN-full (dense) & 72.6 & 96.8 & 53.7 & 87.0 \\ \hline
	\end{tabular}
\end{table}

\begin{figure}[tbp]
	\centering
	\vspace{-0.5cm}
	\subfloat[]{\includegraphics[width=.25\textwidth]{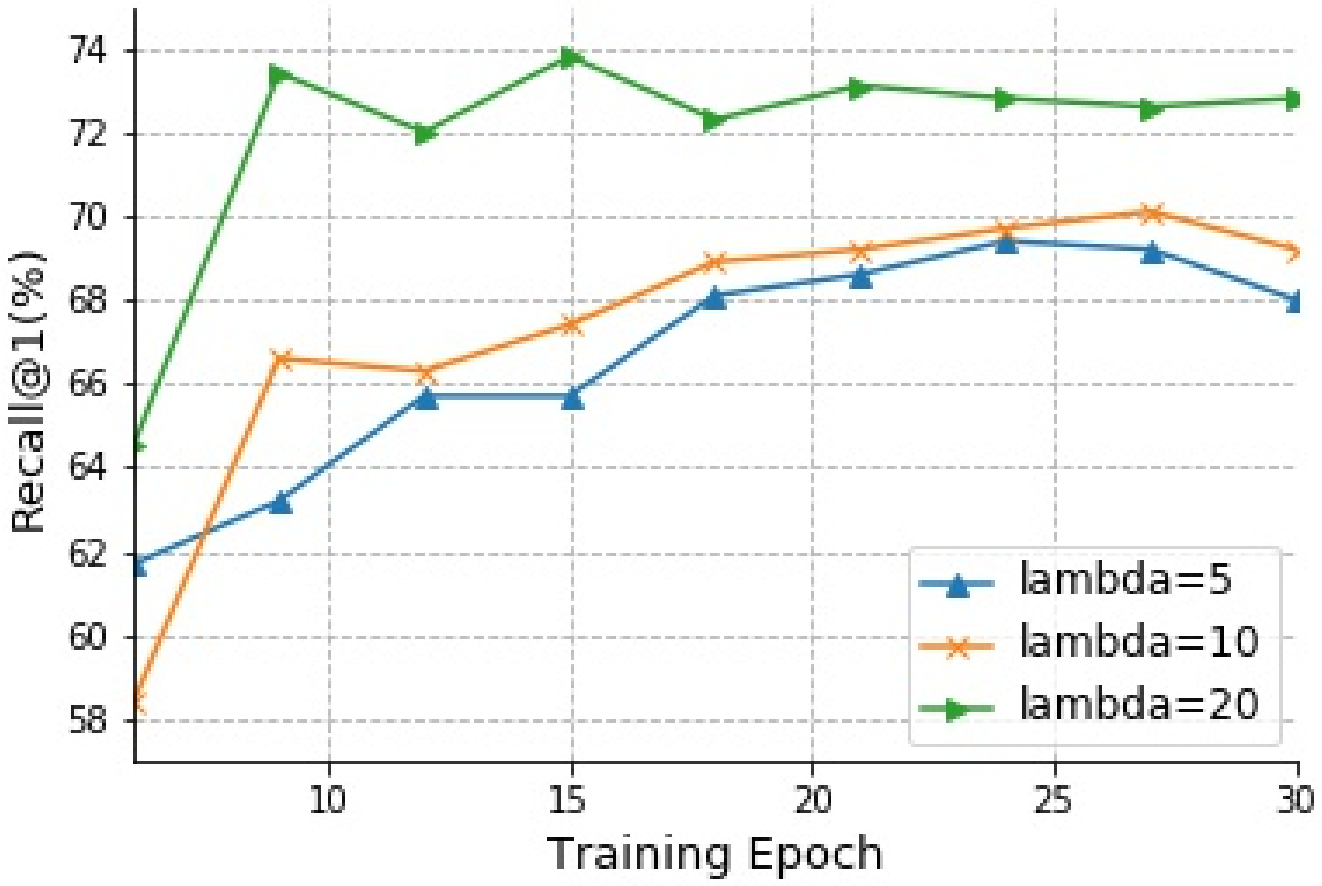}%
		\label{a}}
	\subfloat[]{\includegraphics[width=.25\textwidth]{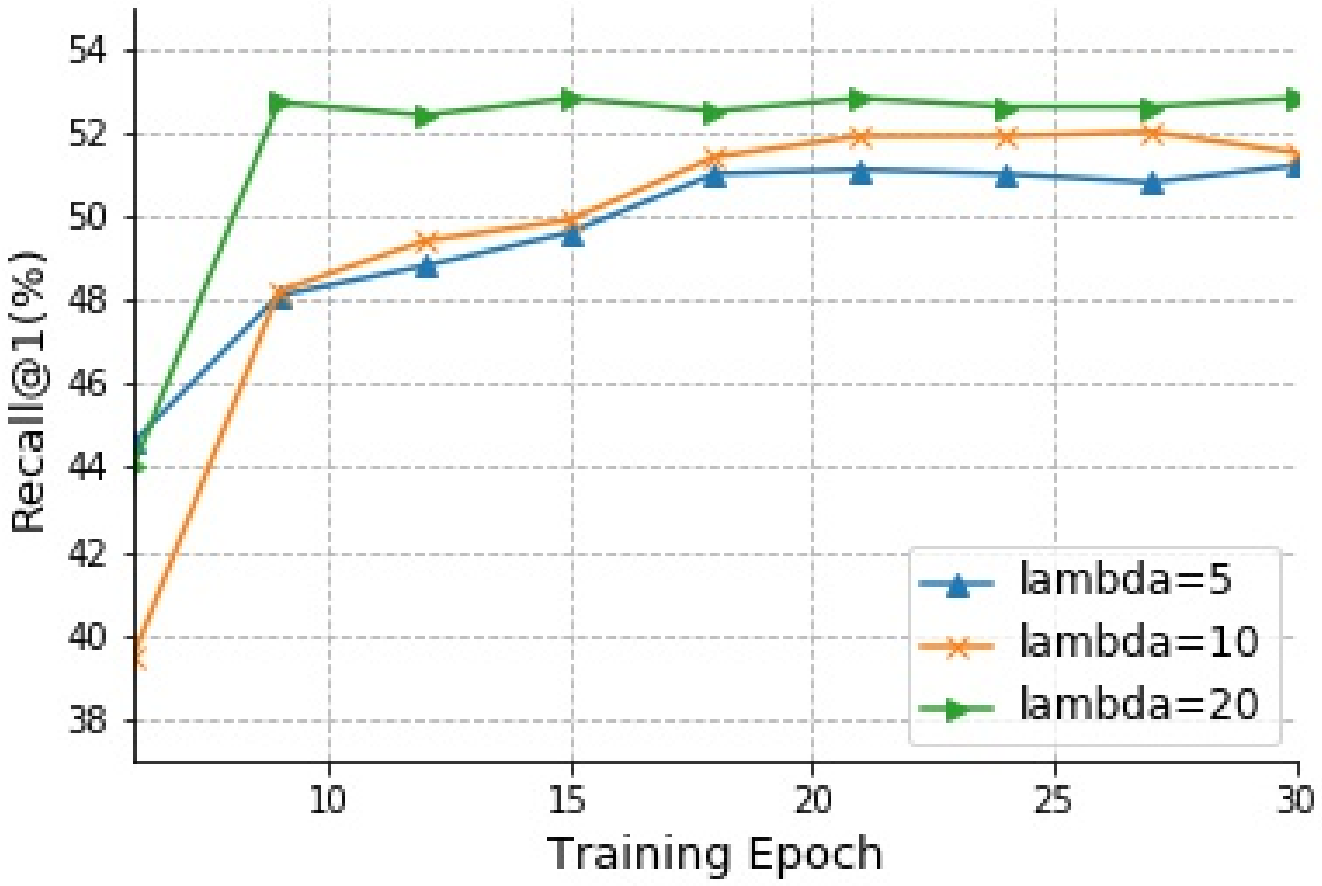}%
		\label{b}}	
	\vspace{-0.4cm} 
	\subfloat[]{\includegraphics[width=.25\textwidth]{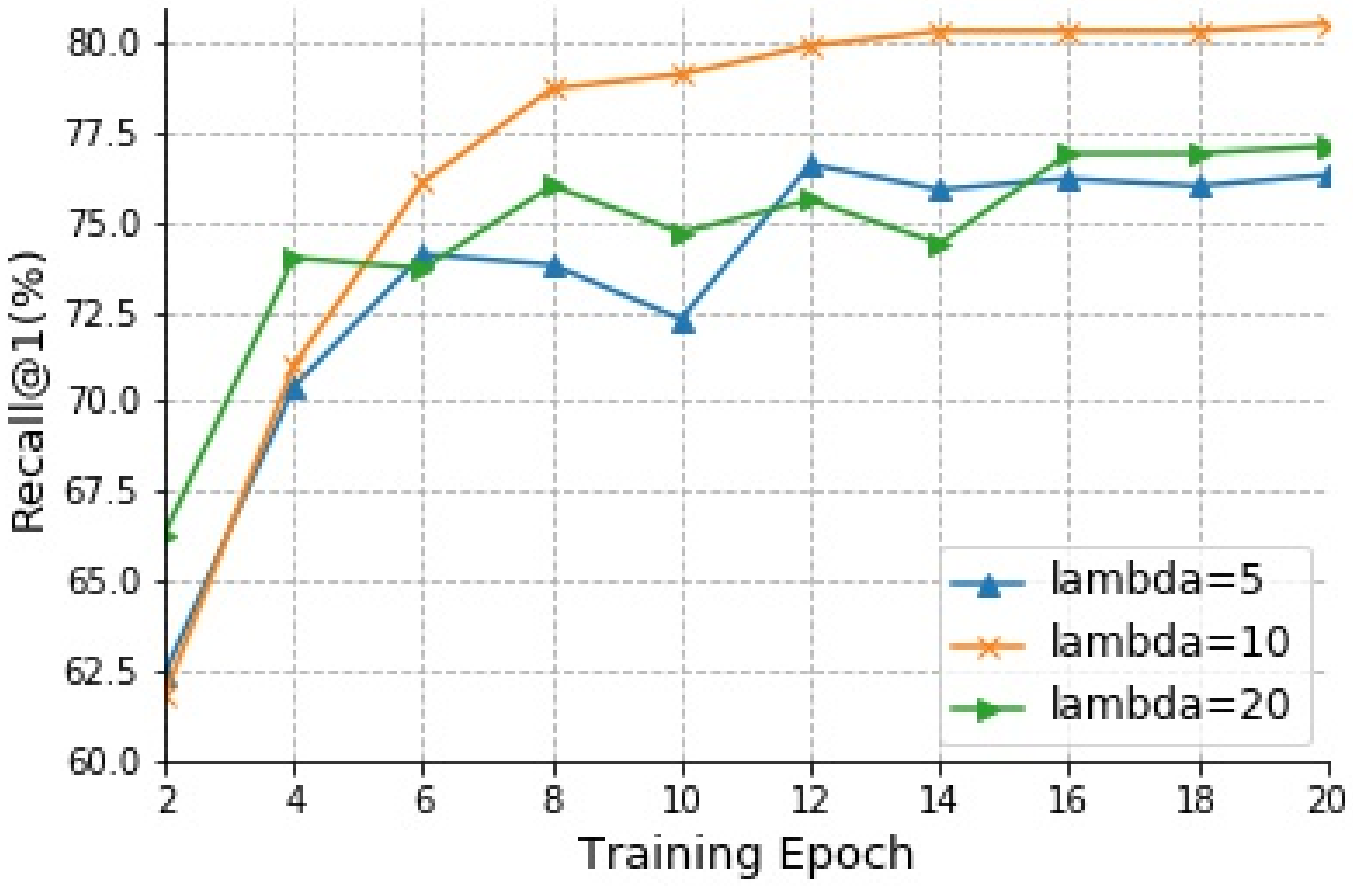}%
		\label{a}}
	\subfloat[]{\includegraphics[width=.25\textwidth]{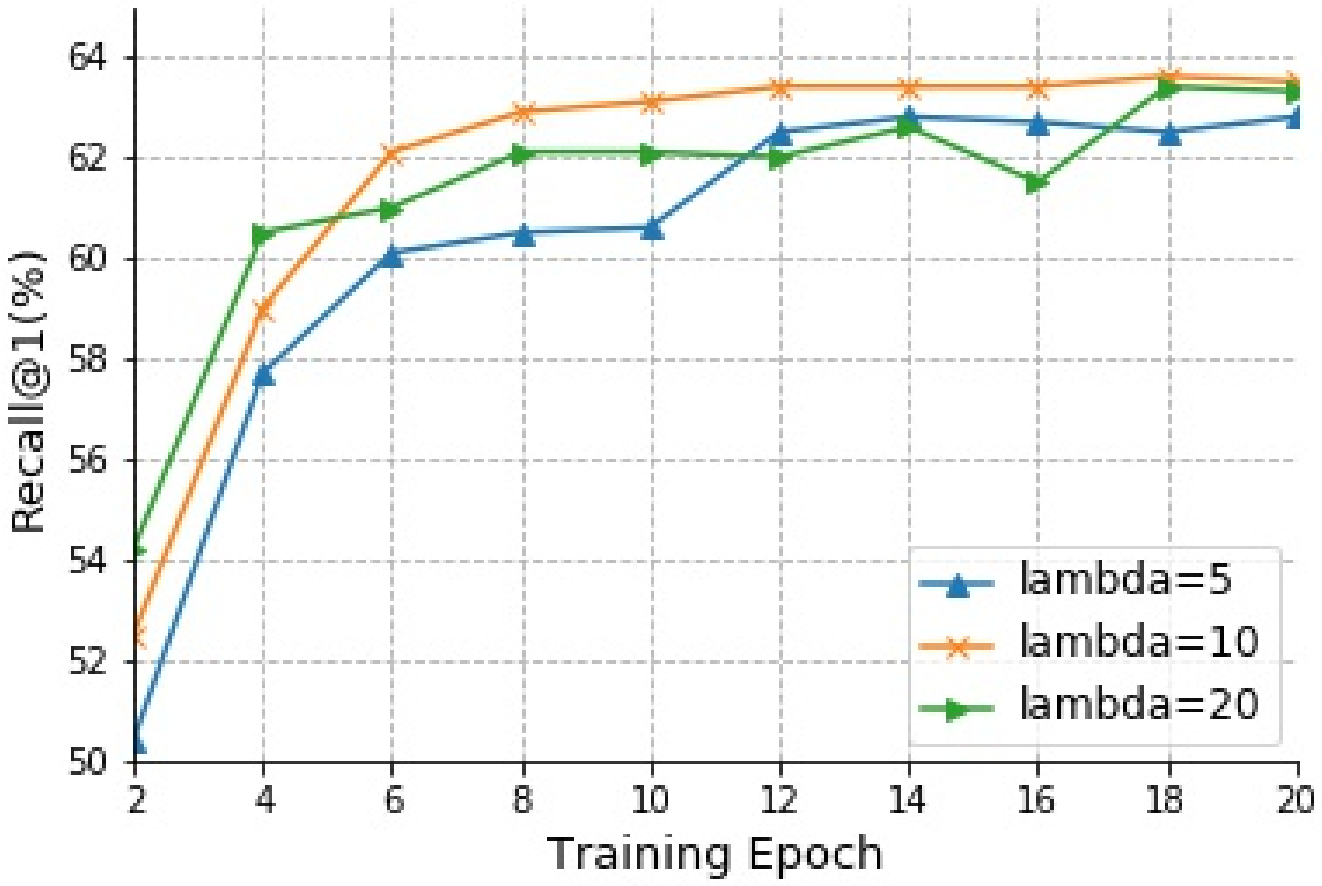}%
		\label{b}}	
	\caption{Comparison of Recall@1 results on Flickr30K and MSCOCO with different $\lambda$ settings. (a) Image-to-text on Flickr30K. (b) Text-to-image on Flickr30K. (c) Image-to-text on MSCOCO. (d) Text-to-image on MSCOCO.}
	\label{fig_sim}
\end{figure}

\noindent
\textbf{Quantitative Analysis.} We provide two versions of our approach, one models the text as a sparse graph and another one models it as a dense graph. We ensemble them by averaging their similarity of image-text pairs, and find that can greatly improve the performance. Note that state-of-the-art methods \cite{wang2019position, lee2018stacked, liu2019focus} also use ensemble model. As shown in Table 1, we can observe that the proposed network outperforms state-of-the-arts with respect to all the evaluation metrics on Flickr30K. Compared with the state-of-the-art method PFAN \cite{wang2019position} that also utilizing the position information of salient regions,  our approach obtains relative R@1 gains with 6.4\% and 7\% at image-to-text and text-to-image matching. Differs from PFAN \cite{wang2019position} that embeds position information into visual representation, our approach employs it as the weight of graph edge. The improvement indicates that structured models object, relation and attribute can greatly improve the matching performance. Although a previous approach SCAN \cite{lee2018stacked} uses similar method to learn object correspondence, our approach achieves more improvement, with nearly 10\% R@1 gain, since it ignores to explicitly learn correspondence of the relation and attribute. In addition, our single model also outperforms their ensemble model by a large margin, and the dense model is better than sparse one as it can discover latent dependencies.

The quantitative results on a larger and more complicated dataset MSCOCO is shown at Table 2. We can observe that our approach can outperform state-of-the-art methods with nearly 2\% improvement in terms of Recall@1, which is more concerned by users in real applications. Our Recall@10 in image-to-text matching is slightly lower than PFAN since noise exists. Compared with SCAN that is most relevant to our work, we suppress it in terms of all the evaluation metrics, getting over 5.5\% and 4.5\% relative Recall@1 improvements on two directions. Note that the sparse model performs better than the dense model, it mainly arises from the sentence in this dataset is more complicated, and thus might incorrectly correlate totally irrelevant words if a fully-connected graph is built.

\subsubsection{Impact of different network structures}
To validate the impact of different network structures, we conduct ablation studies incrementally on Flickr30K. We compare the full dense model and full sparse model with five models: (1) GSMN-w/o graph, which only performs node-level matching. (2) GSMN-w/o i2t, which only applies the node-level matching and structure-level matching on image-to-text direction. (3) GSMN-w/o t2i, which only applies the node-level matching and structure-level matching on text-to-image direction. (4) GSMN-2GCN, its depth of GCN layer is set as 2. (5) GSMN-GRU, a network that only uses GRU instead of Bi-GRU as the text encoder. As shown in Table 3, The two full models outperform all these types of networks, and they largely exceed the network that only performs matching on single direction. Note that GSMN-2GCN requires more computational cost and GPU memory, results show that a deeper network will drop the performance as it additionally considers indirectly connected nodes, which will disturb the learned correspondence. Compared with GSMN-GRU, our approach achieves more improvement on text-to-image graph, it derives from the Bi-GRU can better model the semantic dependency among object, relation and attribute than GRU, and hence the edge weight of textual graph can be accurately reflected. Note that GSMN-w/o i2t gets better performance than GSMN-w/o t2i, because the implicit relation among regions is difficult to be discovered.
\begin{figure*}[htbp]
	\setlength{\abovecaptionskip}{0pt} 
	\setlength{\belowcaptionskip}{-10pt} 
	\begin{center}
		\includegraphics[width=1\linewidth]{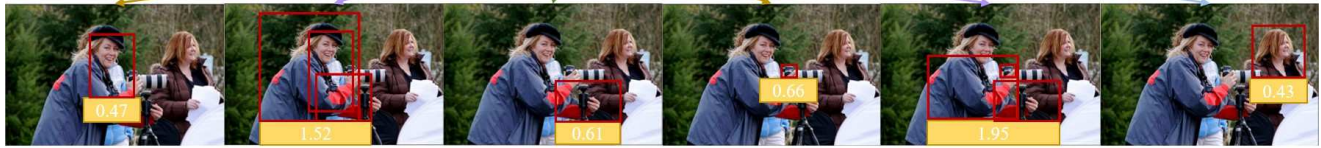}
	\end{center}
	\vspace{-0.3cm}
	\caption{Visualization of node correspondence and phrase correspondence with score inside the box. Best viewed in color.}
\end{figure*}

\begin{figure}[!t]
	\setlength{\abovecaptionskip}{0pt} 
	\setlength{\belowcaptionskip}{0pt} 
	\begin{center}
		\includegraphics[width=1\linewidth]{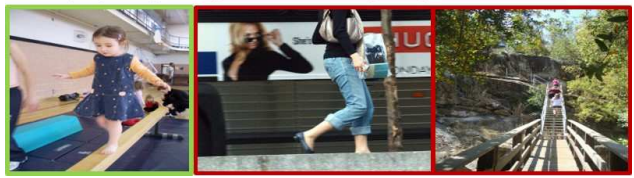}
	\end{center}
	\caption{Visualization of text-to-image matching on Flickr30K. For each text query, we show top 3 ranked images from left to right, where mismatched images are with red boxes and matched images are with green boxes.}
\end{figure}

\subsubsection{Impact of different parameters}
To validate the impact of different parameters, we conduct extensive experiments on two benchmarks. In this work, the most sensitive parameter is the scaling factor \begin{math}\lambda\end{math} that determines the relative weight of different nodes in node-level matching, and the edge weight of textual graph. A large \begin{math}\lambda\end{math} will filter out extensive nodes, and only preserve little nodes that are highly relevant to the specific node. A small \begin{math}\lambda\end{math} is unable to distinguish relevant nodes from irrelevant ones. Hence, an appropriate parameter is important in our proposed network. Here, we investigate the matching performance with setting the \begin{math}\lambda\end{math} as 5, 10 and 20, see figure 3. We observe the Recall@1 on validation set at each training epoch. The top two subfigures are on Flickr30K, it is obvious that when \begin{math}\lambda=20\end{math}, the proposed network yields better Recall@1 on two matching directions, and there is just little difference when the parameter is set as 5 and 10. The bottom two subfigures are on MSCOCO, showing that \begin{math}\lambda=10\end{math} is much better. The different parameter setting on two datasets might be caused by different data distribution.
\begin{figure}[!t]
	\setlength{\abovecaptionskip}{0pt} 
	\setlength{\belowcaptionskip}{0pt} 
	\begin{center}
		\includegraphics[width=1\linewidth]{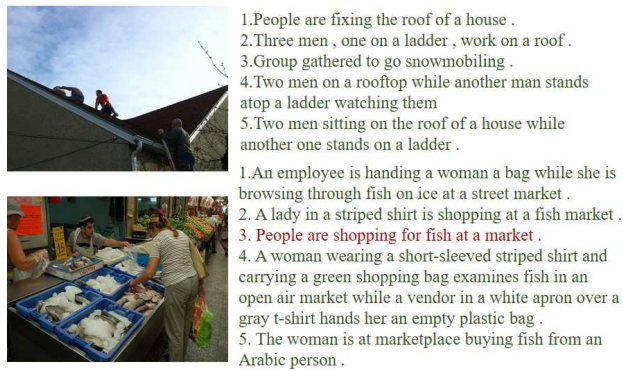}
	\end{center}
	\vspace{-0.3cm}
	\caption{Visualization of image-to-text matching on Flickr30K. For each image query, we show top 5 ranked texts, where mismatched texts are marked as red.}
\end{figure}

\subsubsection{Case Study}
We provide a visualization to show the learned node correspondence and phrase correspondence in Figure 4. Note that we only show the most relevant region for each textual node, it shows different kinds of nodes can associate with their corresponding regions with relatively higher scores. Moreover, we can infer phrase correspondence enclosed by multiple bounding boxes, and their scores are greatly improved. Also, we visualize the text-to-image and image-to-text matching results on Flickr30K, shown in Figure 5 and Figure 6. These show our approach always retrieves the ground truth with a high rank. In addition, our approach is able to learn fine-grained correspondence of the relation and attribute. For example, for the first text query in Figure 5, our network can distinguish different kinds of hats.

\subsection{Conclusion}
In this paper, we propose a graph structured matching network for image-text matching, which performs matching on heterogeneous visual and textual graphs. This is achieved by node-level matching and structure-level matching that infer fine-grained correspondence by propagating node correspondence along the graph edge. Moreover, such a design can learn correspondence of relation and attribute, which are mostly ignored by previous works. With the guidance of relation and attribute, the object correspondence can be greatly improved. Extensive experiments demonstrate the superiority of our network.

\section{Acknowledgements}
This work is supported by the National Natural Science Foundation
of China, Grant No.U19A2057, the Fundamental Research Funds for the Central Universities, Grant No.WK3480000008, the National Key Research and Development Program of China, Grants No.2016QY03D0505, 2016QY03D0503, 2016YFB081304, Strategic Priority Research Program of Chinese Academy of Sciences, Grant No.XDC02040400.

{\small
\bibliographystyle{ieee_fullname}
\bibliography{egbib}
}

\end{document}